\documentclass[preprint,1p,11pt]{elsarticle}
\newcommand{\rev}[1]{\textcolor{black}{#1}}
\usepackage{graphicx}
\usepackage{amssymb}
\usepackage{lineno}
\usepackage{xspace}
\usepackage{amsfonts}
\usepackage{amsmath}
\usepackage{graphicx,subcaption,float}
\usepackage{enumerate}
\usepackage{caption}
\usepackage[table, xcdraw]{xcolor}
\usepackage{booktabs}
\usepackage{algorithm}
\usepackage[noend]{algpseudocode}
\usepackage{hhline}
\usepackage{multirow}
\usepackage{tabularx}
\usepackage{hyperref,url,xspace}
\usepackage{arydshln}
\newcommand{\latinphrase}[1]{\textit{#1}}

\newcommand{\ie}{\latinphrase{i.e.}\xspace}
\newcommand{\eg}{\latinphrase{e.g.}\xspace}

\DeclareMathOperator*{\argmin}{\arg\!\min}

\usepackage{algpseudocode}
\algdef{SE}[DOWHILE]{Do}{doWhile}{\algorithmicdo}[1]{\algorithmicwhile\ #1}
\makeatletter

\makeatletter
\def\@author#1{\g@addto@macro\elsauthors{\normalsize%
    \def\baselinestretch{1}%
    \upshape\authorsep#1\unskip\textsuperscript{%
      \ifx\@fnmark\@empty\else\unskip\sep\@fnmark\let\sep=,\fi
      \ifx\@corref\@empty\else\unskip\sep\@corref\let\sep=,\fi
      }%
    \def\authorsep{\unskip,\space}%
    \global\let\@fnmark\@empty
    \global\let\@corref\@empty  
    \global\let\sep\@empty}%
    \@eadauthor={#1}
}

\def\ps@pprintTitle{%
   \let\@oddhead\@empty
   \let\@evenhead\@empty
   \def\@oddfoot{\reset@font\hfil\thepage\hfil}
   \let\@evenfoot\@oddfoot
}
\makeatother

\begin{document}
\begin{frontmatter}
\title{Privacy and Utility Preserving Sensor-Data Transformations\footnote{Accepted to appear in Pervasive and Mobile computing (PMC) Journal, Elsevier.}}
\author{\mbox{Mohammad~Malekzadeh$^{*}$, Richard G. Clegg$^{*}$, Andrea Cavallaro$^{*}$, Hamed  Haddadi$^{**}$}\\
\mbox{\{m.malekzadeh, r.clegg,  a.cavallaro\}@qmul.ac.uk, h.haddadi@imperial.ac.uk}\\
{$^{*}$Queen Mary University of London, $^{**}$Imperial College London\\
  }
}
\date{}
\begin{abstract}
Sensitive inferences and user re-identification are major threats to privacy when raw sensor data from wearable or portable devices are shared with cloud-assisted applications. To mitigate these threats, we propose mechanisms to transform sensor data before sharing them with  applications running on users' devices. These  transformations aim at eliminating patterns that can be used for user re-identification or for inferring potentially sensitive activities, while introducing a minor utility loss for the target application (or task).  We show that, on gesture and activity recognition tasks, we can  prevent inference of potentially sensitive activities while keeping the reduction in recognition accuracy of non-sensitive activities to less than 5 percentage points. We also show that we can  reduce the accuracy of user re-identification and of the  potential inference of gender to the level of a random guess, while keeping the accuracy of activity recognition  comparable to that obtained on the original data.
\end{abstract}

\end{frontmatter}

\section{Introduction}
\label{sec:intro}

Sensors such as accelerometer, gyroscope, and magnetometer, embedded in personal smart devices generate data that can be used to monitor users' activities, interactions, and mood~\cite{katevas2015walking, hansel2018potential, irfan2018anomaly}. Applications~(apps) installed on smart devices can get access to raw sensor data to make {\em required} (\ie~desired) inferences for tasks such as gesture or activity recognition. However, sensor data can also facilitate some potentially {\em sensitive} (\ie~undesired) inferences that a user might wish to keep private, such as discovering smoking habits~\cite{scholl2012feasibility} or revealing personal attributes such as age and gender~\cite{riaz2015one}. Some patterns in raw sensor data may also enable user re-identification~\cite{neverova2016learning}.

Information privacy can be defined as {\em ``the right to select what personal information about me is known to what people''}~\cite{westin1968privacy}. To preserve privacy, we need mechanisms to control the type and amount of information that providers of  cloud-assisted apps can discover from sensor data. The main objective is to move from the current binary setting of granting or not sensor permission to an app, toward a model that allows users to grant each app permission over a controlled range of inferences according to the target task. 
The challenging task is to design a mechanism with an acceptable trade-off between the protection of {\em sensitive} information and the maintenance of the {\em required} information for an inference~\cite{ghosh2012universally}. To this end, we use {\em neutral} inferences that are irrelevant to the target task and not critical to the user's privacy.

\begin{figure}[t]
	\centering
	\includegraphics[width=\textwidth]{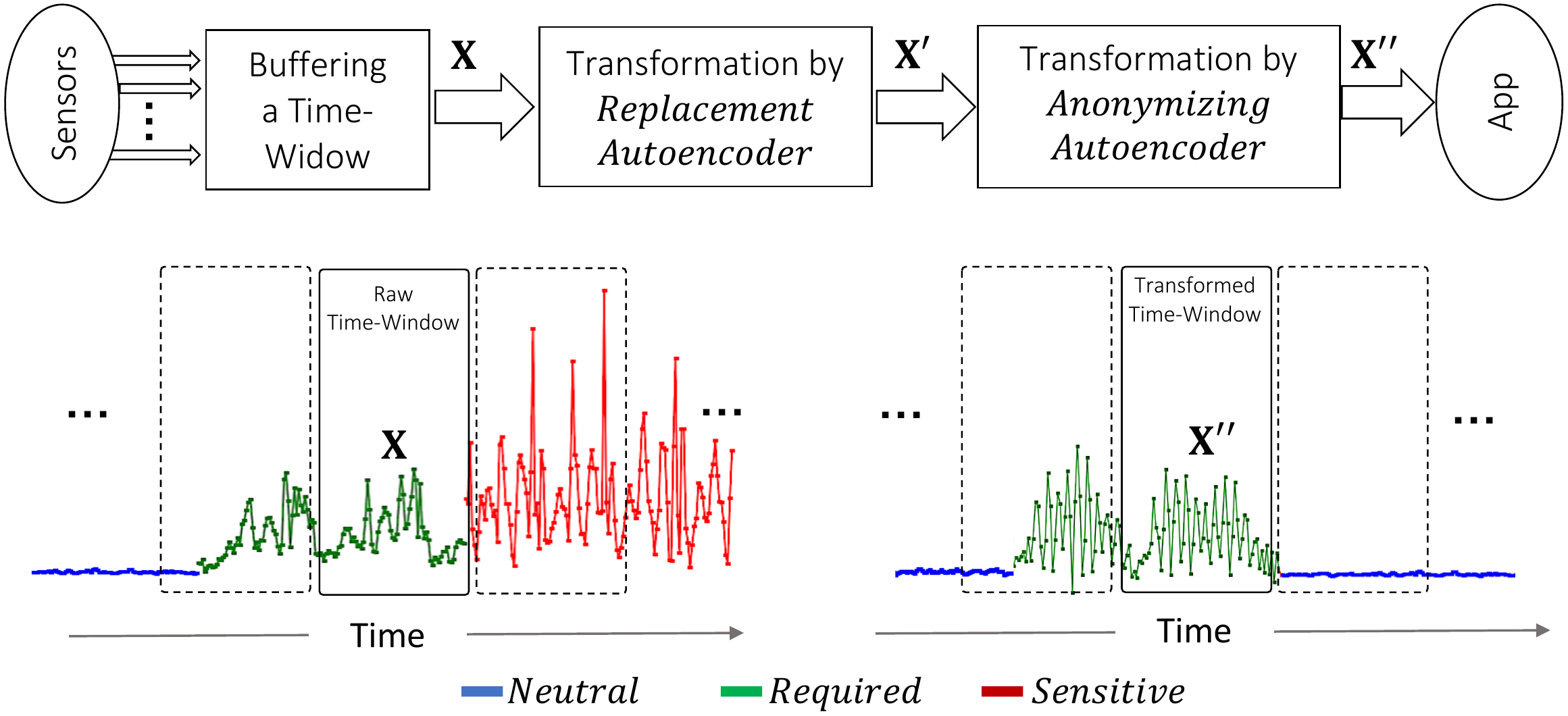}
	\caption{ (Top) the data flows in the compound framework. At the test-time, first RAE automatically replaces {\em sensitive} time-windows
		with non-sensitive {\em neutral} data,
		while {\em required} time-windows
		are passed with minimal distortions. Then, AEE transforms data to reduce the chance of {\em user re-identification}. (Bottom) visual illustration of our transformation mechanism. Depicted signals show accelerometer data transformation for standing, walking, and jogging activities respectively as {\em neutral}, {\em required}, and {\em sensitive} inferences (from experiment of Section~\ref{sec:evl_comb}).} 
	\label{fig:overview}
\end{figure}

As specific example of categorization of  {\em sensitive}, {\em required}, and {\em neutral} information, let us consider a smartwatch step-counter app: {\em required} information is essential for the app's utility, such as walking or stair stepping; {\em sensitive} information is about activities a user wishes to keep private, such as smoking or typing on a keyboard, or information such as gender; whereas {\em neutral} information leads to inferences that are neither {\em required} nor {\em sensitive}, such as when the user sits or stands. Note that two types of information (\ie~{\em required} and {\em sensitive}, or {\em neutral} and {\em sensitive}) are sometimes entangled in the data of the same temporal window of sensor measurements. While locally differentially private mechanisms~\cite{dwork2014algorithmic} provide plausible deniability guarantees when  estimating, for example, the mean or frequency of a common variable among users~\cite{bittau2017prochlo},  with multi-dimensional data  released sequentially a more practical privacy model  is {\em inferential privacy}~\cite{menasria2018purpose, huang2017context, edwards2015censoring} that measures the difference between an adversary's belief about {\em sensitive} inferences before and after observing the released data. 

We assume the app provider is honest in stating its {\em required} inferences that need to be made on the data, but it is also  curious about making other unstated inferences that may reveal sensitive information and thus violate privacy. We define  {\em utility} as the accuracy in making the {\em required} inferences on the released data, and {\em privacy loss} as the accuracy in making {\em sensitive} inferences. We define the app provider as adversary and quantify the {\em privacy loss} as the improvement in the adversary's posterior belief on making a {\em sensitive} inference by observing the data. Our proposed mechanisms aim to minimize the {\em privacy loss} while maintaining the {\em utility} of the raw data.

In this paper, we  present mechanisms for transforming time-windows of sensor data to preserve privacy and utility during  information disclosure~\cite{miklau2007formal, du2012privacy, ghosh2016inferential} to a {\em honest-but-curious} app running on users' devices.
 Specifically, we introduce a Replacement AutoEncoder~(RAE) to protect {\em sensitive} inferences and an Anonymizing AutoEncoder~(AAE) to prevent {\em user re-identification}, as well as a compound architecture by cascading the RAE and AAE (see Figure~\ref{fig:overview}). The RAE and AAE can be deployed as interface into the devices' operating system to enable users to choose whether to share their sensor data with an app directly or after transformations.  To validate our mechanisms, in addition to using available datasets, we collected a dataset of activity recognition using smartphone sensors, which is made publicly available\footnote{Code and data are available at: \url{https://github.com/mmalekzadeh/motion-sense}}.
Experiments on gesture and activity recognition show that the RAE substantially reduces the {\em  privacy loss} for {\em sensitive} gestures or activities while limiting the reduction in the {\em utility} of the {\em required} and {\em neutral} gestures or activities to less than 5 percentage points.  
Furthermore,  results on an activity recognition dataset of 24 users show a promising trade-off, with the {utility} maintained over 92\% and a reduction of the {\em privacy loss} in user re-identification to less than 7\%~, from an initial 96\% on the raw data. We also show that our mechanisms lead to models that can generalize across datasets and can be applied to new data of unseen users.

\section{Related Work}

Privacy-preserving mechanisms for time-series data can be implemented through {\em perturbations}, {\em synthesis}, {\em filtering}, or {\em transformations}. 

Mechanisms using {\em perturbations} hide sensitive patterns by adding a crafted noise to each time-window of the time-series. The objective is to prevent perturbed data from including sufficient information to accurately reconstruct the original data~\cite{amar2018information}. Because an independent and identically distributed noise can be easily removed from correlated time-series~\cite{wang2017cts}, to reduce the risk of information leakage, the correlation between noise and original time-series should be indistinguishable~\cite{zhu2015correlated}. For multi-dimensional sensor data, it is not easy to find a reliable model of correlation between the original data and an adequate noise. Hence, when general time-series perturbation approaches are extended to sensor data, effectively hiding sensitive patterns without excessively perturbing the non-sensitive ones is very challenging.  

Data can also be {\em synthesized} to maintain some  required statistics of the original data without information that can be used for re-identification. Adversarial learning enables one to approximate an underlying distribution to generate new data that are similar to the existing ones~\cite{kingma2014auto, goodfellow2014generative}. To provide a privacy guarantee, generators can be trained under the constraint of differential privacy~\cite{beaulieu2017privacy, acs2018differentially} or with constraints on the type of information that should be unsynthesized in the data~\cite{laforet2015individual}. However, these mechanisms are used for offline dataset publishing by a data aggregator~\cite{esteban2017real}, not for online data transformation at the user side.

{\em Filtering} can be used to remove unwanted components only in temporal intervals that include sensitive information.  MaskIt~\cite{gotz2012maskit} releases location time-series when users are at a regular workplace and suppresses them when they are in a {sensitive} place, such as a hospital. A Markov chain built on a pre-defined set of conditions is employed for each user. A Dynamic Bayesian  Network model can be used offline to replace {sensitive} time-windows that indicate users' stress, while keeping non-sensitive time-windows corresponding to their walking periods~\cite{saleheen2016msieve}. 

\begin{table}[t!]
	\centering
	\footnotesize
	\caption{
		Privacy-preserving mechanisms for sharing time series. Key - {\em Local}: applied on the user side (instead of being done globally by a data curator); {\em GesAct}: hides users' {sensitive} gestures or activities; {\em Identity}: prevents user {re-identification}; {\em Sensors}: evaluated on sensor data; {\em Unseen}: can be used for data of users who did not contribute training data.}
	\label{tab:rwcomp}
	\resizebox{\textwidth}{!}{%
		\begin{tabular}{llccccc}
			{\em  Mechanism} & {\em  Reference} & {\em  Local} & {\em  GesAct} & {\em  Identity} & {\em  Sensor} & {\em  Unseen} \\
			\hline 
			Perturbations & \cite{ amar2018information, wang2017cts, zhu2015correlated} &    & $\checkmark$ &   & $\checkmark$  & \\
			\arrayrulecolor{black!20}\cline{1-7}\arrayrulecolor{black!100}
			Synthesis & \cite{ beaulieu2017privacy,laforet2015individual,esteban2017real} & &  & $\checkmark$ &   & \\
			\arrayrulecolor{black!20}\cline{1-7}\arrayrulecolor{black!100}
			\multirow{2}{*}{Filtering}& \cite{gotz2012maskit, shamsabadi2018distributed, psychoula2018deep} &  $\checkmark$ &   $\checkmark$ &   &  &  \\
			\arrayrulecolor{black!0}\cline{2-7}\arrayrulecolor{black!100}
			& \cite{saleheen2016msieve} & &    $\checkmark$ &  $\checkmark$  & $\checkmark$ &  \\
			\arrayrulecolor{black!20}\cline{1-7}\arrayrulecolor{black!100}
			\multirow{4}{*}{Transformations}&
			\cite{huang2017context,edwards2015censoring} & $\checkmark$  &    &  $\checkmark$ & &    \\
			\arrayrulecolor{black!0}\cline{2-7}\arrayrulecolor{black!100} 
			&{\cite{menasria2018purpose,lu2017information} }& &   & $\checkmark$  & $\checkmark$ &  \\
			\arrayrulecolor{black!0}\cline{2-7}\arrayrulecolor{black!100}
			& \cite{raval18olympus} & $\checkmark$   &   & $\checkmark$ & $\checkmark$ & $\checkmark$ \\
			\arrayrulecolor{black!20}\cline{1-7}\arrayrulecolor{black!100}
			\textbf{Filtering \& Transformations} & {\bf Ours} & $\checkmark$   & $\checkmark$ & $\checkmark$ & $\checkmark$ & $\checkmark$ \\ \hline
		\end{tabular}
	}
\end{table}


{\em Transformations} can reduce the amount of {sensitive} information in the data by reconstruction~\cite{huang2017context} or by projecting each data sample into a lower dimensional latent representation~\cite{edwards2015censoring,osia2018deep}.
The information bottleneck in the hidden layers of neural networks helps to capture the main factors of variation in the data and to identify and obscure {sensitive} patterns in the latent representation~\cite{edwards2015censoring}, as well as during the reconstruction from the extracted low-dimensional representation~\cite{raval18olympus, shamsabadi2018distributed}. Global mechanisms involve a trusted data curator and, based on the information bottleneck principle, compress sensor data to reduce sensitive information that is irrelevant to the main task~\cite{menasria2018purpose}.

Table~\ref{tab:rwcomp} compares methods related to our work. A privacy-preserving mechanism can be run globally or locally. Global mechanisms involve a trusted data curator that has access to the original data and offer a data transformation service to remove sensitive information before data publishing~\cite{menasria2018purpose, lu2017information, xiao2018information}. Local mechanisms, instead, manipulate data at the user side, without relying on a trusted curator~\cite{gotz2012maskit,raval18olympus, shamsabadi2018distributed}. Our mechanisms run {\em locally} and can be used by users who did not contribute training data ({\em unseen} users). 

\section{Sensor-Data Transformation}
We first introduce the Replacement AutoEncoder~(RAE) that protects {\em sensitive} inferences, then we present the Anonymizing AutoEncoder~(AAE) that prevents {\em user re-identification}. The notations we use in this paper are shown in Table~\ref{tab:toc}.

\begin{table}[t!]
\centering

\footnotesize
\caption{\rev{Main notation used in this paper.}}\label{tab:toc}
\centering 
\resizebox{\textwidth}{!}{%
\begin{tabular}{p{2.5cm} p{10.4cm}}
\toprule
$\mathbf{x}_{{sj}}\in \mathbb{R}$ & reading from sensor component ${s}$ at sampling instant ${j}$; 
\\
$\mathbf{X}\in \mathbb{R}^{M\times W}$ &  time-window of $W$ samples from $M$ sensors;
\\
$\mathcal{X}^{i}, \mathcal{X}^{o} \subseteq \mathbb{R}^{M\times W}$ & input and output datasets, respectively, for training the RAE;
\\
$\mathbf{X'}$, $\mathbf{X''} \in \mathbb{R}^{M\times W}$ & output of the RAE and the AAE, respectively;
\\
$\mathbf{U} \in \{0,1\}^{N}$ & $N$-dim vector representing the identity of a user ($\sum_{i=1}^{N}u_i =1$);
\\
$\mathbf{Y}\in \{0,1\}^{B}$ & $B$-dim vector representing a gesture/activity ($\sum_{i=1}^{B} y_i =1$);
\\
$\mathrm{I}(\cdot;\cdot)$ & mutual information function;
\\
$\mathrm{d}(\cdot,\cdot)$  & distance function between two time-series (\eg~Mean Squared Error). 
\\
\bottomrule
\end{tabular}
}
\end{table}

\subsection{Replacement AutoEncoder} \label{sec:rae}
Deep neural networks~(DNNs) are powerful machine learning algorithms that progressively learn hierarchical and relevant representations of their training data. Earlier layers of a DNN can encode generic low-level data patterns and later layers can capture more specific high-level features. Autoencoders learn features from data through minimizing the differences (\eg~mean squared error or cross entropy) between the input and its reconstruction. The information bottleneck~\cite{bengio2013representation} in the hidden layers forces an autoencoder to put more attention on the descriptive data patterns in order to generalize the model. 

Let a fixed-length time-window of sensor data, $\mathbf{X}=(\mathbf{x}_{sj})\in \mathbb{R}^{M\times W}$, contain some specific patterns that are utilized to recognize the gesture or activity of the user at that specific time-point. For example, let us consider an smartwatch app which counts users' daily steps. Users may want this app to only be able to infer activities that are {\em required} for step counting task, not other activities such as smoking or eating that may be considered {\em sensitive}. The main idea of RAE is to automatically recognize and replace each time-widows that reveals {\em sensitive} activities with a same dimension data that simulates a {\em neutral} activity, such as standing or sitting, which does not affect the step counter utility.

\begin{figure}[t!]

	\centering
	\includegraphics[width=.9\linewidth]{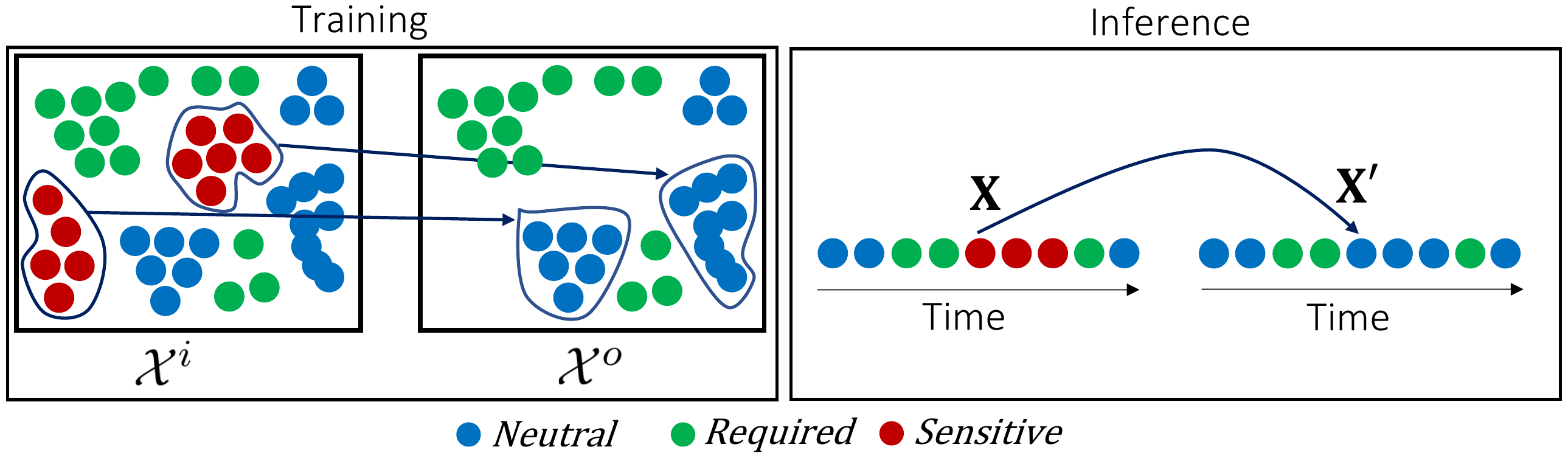} 
	\caption{
	\label{rae} Circles represent time-windows in the input ($\mathcal{X}^i$) and output ($\mathcal{X}^o$) datasets for training the RAE. We first make a copy of the original input dataset and replace every {\em sensitive} time-window with a randomly chosen {\em neutral} one to prepare the output dataset for training the RAE. Then, the RAE is trained to transform each $\mathbf{X}^{i}$ to the corresponding $\mathbf{X}^{o}$. At inference time, RAE can replace unseen {\em sensitive} time-windows with data that simulates {\em neutral} ones.}
\end{figure}

Let the training dataset include labeled sample time-windows, each belonging to one of the following categories: {\em required}, {\em sensitive}, or {\em neutral}. Let $\mathcal{X}^{i}$ be the {\em input} dataset and $\mathcal{X}^{o}$  be the {\em output} dataset, with  a one-to-one relationship between each $\mathbf{X}^{i} \in \mathcal{X}^{i}$ and an $\mathbf{X}^{o}  \in \mathcal{X}^{o}$ explained in Figure~\ref{rae}. Basically, data samples of {\em sensitive} classes in $\mathcal{X}^{i}$ are randomly replaced with data samples from one of the {\em neutral} classes to build $\mathcal{X}^{o}$. Therefore  $\mathcal{X}^{o}$ contains only samples from the {\em required} and {\em neutral} classes.  The RAE is then trained to transform each $\mathbf{X}^{i}$ to the corresponding $\mathbf{X}^{o}$, subject to a loss function, $\mathrm{L_R}(\mathbf{X}^{i}, \mathbf{X}^{o})$, which  calculates the difference between the input of the RAE and its corresponding output.  

Let a replacement be defined as {\em privacy-preserving} if its outcome removes, or practically bounds, the possibility of revealing {\em sensitive} inferences. If $\mathbf{X^r}$ is a {\em privacy-preserving} replacement for {\em sensitive} data $\mathbf{X}$, the RAE aims to implement the following operation:
\begin{equation} \label{pwfunc}
\mathbf{X'}
= RAE(\mathbf{X)} =
\begin{cases}
\mathbf{X^r} & \text{if $\mathbf{X}$ reveals a sensitive inference,} \\
\mathbf{X} & \text{otherwise,} \\
\end{cases}
\end{equation}
where the {\em privacy loss} of the replaced data, $\mathbf{X'}$, is equivalent to the amount of sensitive information it reveals. If $\mathrm{R}(\mathbf{X ; \theta})$ is an autoencoder with parameter set $\theta \in \Theta$, and $\mathrm{L_R}(\cdot, \cdot)$ is the autoencoder's loss function, we define the optimal parameter set for the RAE as  
\begin{equation} \label{eq:rae}
\theta^* = \argmin_{\theta \in \Theta}  \mathrm{L_R}\left(\mathrm{R}\left(\mathcal{X}^{i} ; \theta\right), \mathcal{X}^{o}\right),
\end{equation}
which can be achieved through a neural network optimization process~\cite{ malekzadeh2018replacement}. The implementation details of RAE are explained in Section~\ref{sec:evl}. 
\subsection{Anonymization AutoEncoder}\label{sec:anon}

As time-windows that do not reveal sensitive inferences are supposed to pass through the RAE with minimum distortion, they may be used for other malicious purposes such as user re-identification. As a motivational example, consider participants in a study for a new treatment who share their daily sensor data with researchers~\cite{rodriguez2017waist}. These participants may want to minimize the risk of being re-identified by those who will access their released data. Therefore, their sensor data should be released in a way that the required information for the medical study, such as patients’ daily activities, can be accurately inferred, while other motion patterns that facilitate user re-identification are obscured.

We define the  data with the user's identifiable information obscured as the anonymized sensor data, $\mathbf{X''}$.
Considering $\mathrm{A}(\cdot)$ as a potential data transformation function and  $\mathbf{X}$ the data we want to anonymize, we define the fitness function $\mathrm{F}(.)$  as 

\begin{equation}\label{eq:aae}
    \mathrm{F}\left(\mathrm{A}\left(\mathbf{X}\right)\right)
    =
    \beta_{i}
    \mathrm{I}\left(\mathbf{U};\mathrm{A}\left(\mathbf{X}\right)\right)
    -
    \beta_{a}\mathrm{I}\left(\mathbf{Y}; \mathrm{A}\left(\mathbf{X}\right)\right)
    +
    \beta_{d} \mathrm{d}\left(\mathbf{X},\mathrm{A}\left(\mathbf{X}\right)\right),
\end{equation}
where the non-negative, real-valued weights $\beta_i$, $\beta_a$ and $\beta_d$ determine the trade-off between privacy loss and utility. As it is discussed in Section~\ref{sec:evl}, the desired trade-off is established through cross validation over the training dataset. 

Let the anonymization function,  $\mathcal{A}(\cdot)$, that transforms $\mathbf{X}$ into  $\mathbf{X''}$ be 
\begin{equation}\label{eq:anon}
    \mathcal{A}\left(\mathbf{X}\right) = \argmin_{\mathrm{A}\left(\mathbf{X}\right)}{ \mathrm{F}\left(\mathrm{A}\left(\mathbf{X}\right)\right)}.
\end{equation}
The threefold objective of Eq.~(\ref{eq:aae}) is to minimize $\mathrm{I}\left(\mathbf{U};\mathcal{A}\left(\mathbf{X}\right)\right)$, the mutual information between the random variable that specifies the identity of current user and the anonymized  data; to maximize $\mathrm{I}\left(\mathbf{Y}; \mathrm{A}\left(\mathbf{X}\right)\right)$, the mutual information between the random variable that captures the user activity and the anonymized data (\ie~to minimize its negative value); and, to avoid large data distortions by minimizing $\mathrm{d}\left(\mathbf{X},\mathrm{A}\left(\mathbf{X}\right)\right)$, the distance between raw and anonymized data.

As we cannot practically search over all possible transformation functions,  we consider a DNN and look for the optimal parameter set through training. To approximate the required mutual information terms, we reformulate the optimization problem in Eq.~(\ref{eq:anon}) as a DNN optimization problem.
Let $\mathrm{A}(\mathbf{X}; \theta)$ be a DNN, where  $\theta$ is the  parameter set of the DNN.
The network {optimizer} finds the optimal parameter set $\theta^{*}$ by searching the space of all possible parameter sets, $\Theta$, as:
\begin{equation}\label{eq:id_act}
\theta^{*}=\argmin_{ \theta \in \Theta} \beta_i\mathrm{I}\left(\mathbf{U};\mathrm{A}\left(\mathbf{X}; \theta\right)\right)
-
\beta_a\mathrm{I}\left(\mathbf{Y}; \mathrm{A}\left(\mathbf{X}; \theta\right)\right)
+
\beta_d\mathrm{d}\left(\mathbf{X}, \mathrm{A}\left(\mathbf{X}; \theta\right)\right),
\end{equation}
where $\mathcal{A}(\cdot; \theta^{*})$ is the optimal data anonymizer for a general $\mathcal{A}(\cdot)$ in Eq.~(\ref{eq:anon}). Again, we can obtain $\theta^{*}$ using a stochastic optimization algorithm~\cite{kingma2014adam}. A key contributor to the AAE training is the following multi-objective loss function, $\mathrm{L_A}$, which implements the fitness function $\mathrm{F}\left(\mathrm{A}\left(\mathbf{x}\right)\right)$ of Eq.~(\ref{eq:anon}):
\begin{equation}\label{eq:ws}
        \mathrm{L_A} = \beta_{i}\mathrm{L}_{i}-\beta_{a}\mathrm{L}_{a}+\beta_{d}\mathrm{L}_{d},
\end{equation}
where $\mathrm{L}_{a}$ and $\mathrm{L}_{d}$ are \textit{utility losses} that can be customized based on the target task requirements, whereas $\mathrm{L}_{i}$ is a \textit{privacy loss} that helps the AAE remove user-specific patterns that facilitate user re-identification.
 
Practically, the categorical cross-entropy loss function for classification, $\mathrm{L}_{a}=\mathbf{Y}\log(\hat{\mathbf{Y}})$, aims to preserve activity-specific patterns,
where $\hat{\mathbf{Y}}$, the output of a softmax function, is a $B$-dimensional vector of probabilities for the prediction of the activity label. To tune the desired privacy-utility trade-off,  the distance function that controls the amount of distortion, $\mathrm{L}_{d}$, forces $\mathbf{X''}$ to be as similar as possible to the input $\mathbf{X}$:
\begin{equation}
        \mathrm{L}_{d} = \frac{1}{M\times W} \sum_{s=1}^{M}\sum_{j=1}^{W} (\mathbf{x}_{{sj}}-\mathbf{x''}_{{sj}})^2.
\end{equation}
Finally, the privacy loss, ${L}_{i}$, the most important term of our multi-objective loss function that aims to minimize sensitive information in the data, is defined as:
\begin{equation} \label{eq:lid}
        \mathrm{L}_{i} =
        -\left(
            \mathbf{U}\cdot \log
            \left(
                \textbf{1}^{N}- \hat{\mathbf{U}}
            \right)
            +
            \log
            \left(
                1-max
                \left(
                    \hat{\mathbf{U}}
                \right)
            \right)
        \right),
\end{equation}
where $N$ is the number of users in the training set, $\textbf{1}^{N}$ is the all-one column vector of length $N$, $\mathbf{U}$ is the true identity label for  $\mathbf{X}$, and $\hat{\mathbf{U}}$ is  the output of the softmax function, the $N$-dimensional vector of probabilities  learned by the classifier (\ie ~the probability of each user label, given the input). $\mathbf{U}\cdot \log(\textbf{1}^{N}- \hat{\mathbf{U}})$ is dot product of row vectors.

The goal of training AAE is to minimize the {\em privacy loss} by minimizing the amount of information leakage from $\mathbf{U}$ to $\mathbf{X''}$. Hence, we use adversarial training to approximate the mutual information by estimating the posterior distribution of the sensitive data given the released data~\cite{malekzadeh2018mobile}.

\section{Evaluation}\label{sec:evl}

To evaluate the RAE, we use \rev{four} benchmark datasets of gesture and activity recognition including at least 10 different labels: Opportunity~\cite{chavarriaga2013opportunity}, Skoda~\cite{zappi2008activity}, Hand-Gesture~\cite{bulling14_csur}, \rev{and Utwente~\cite{shoaib2016complex}}. To evaluate the AAE, we need a dataset containing several users to show how we can hide users' gender or identity. Therefore, we use MotionSense~\cite{malekzadeh2018protecting} that contains the collected data of 24 users in a range of gender, age, and height who performed 6 activities. We also evaluate the compound architecture (RAE+AAE) on a case study using the MotionSense dataset. 

Opportunity~\cite{chavarriaga2013opportunity} is composed of the collected data of 4 users and there are 18 gestures classes. 
Each record in this dataset comprises 113 sensory readings from various types of body-worn sensors like accelerometer, gyroscope, magnetometer, and skin temperature. 

Skoda~\cite{zappi2008activity}  is collected by an assembly-line worker in a car production company wearing 19 accelerometer sensors on his {right} and {left} arm and performing a set of pre-specified experiments.

Hand-Gesture~\cite{bulling14_csur} includes data from accelerometer and gyroscope sensors attached to the upper and lower arm. There are two users performing 12 classes of hand movements. Each record in this dataset has 15 real-valued sensor readings. 

Utwente~\cite{shoaib2016complex} includes the data of 6 participants  performing several activities, including potentially sensitive smoking activity,  while wearing a smart-phone on their wrist. Accelerometer, gyroscope, and magnetometer data are collected.  The whole dataset is publicly available in a single file with activity labels only.

MotionSense~\cite{malekzadeh2018protecting} is collected with a smart-phone kept in the users' front pocket. A total of 24 users performed 6 activities in 15 trials in the same environment and conditions. It  includes acceleration, rotation, gravity, and attitude data. Each record in this dataset includes 12 real-valued sensor readings.  

Table~\ref{tbl_dataset} summarizes the gesture/activity classes of the five datasets. For Opportunity, we use four trials as the training data, and consider the last trial as the testing data. For other datasets, we consider 80\% of the data as the training set and the rest as the testing set. The {\em null} class in the gesture datasets refers to data that cannot be mapped to a known behavior. All the gesture datasets are resampled to 30Hz sampling rate.
\begin{table}[t!]
	\centering
	\caption{ Gesture/Activity classes and properties of each dataset used for evaluation.} \label{tbl_dataset}
	\resizebox{\textwidth}{!}{%
	\begin{tabular}{l|lll|ll}
		\cline{1-6}
		&\multicolumn{3}{c|}{\textbf{Gesture Datasets}} &\multicolumn{2}{c}{\textbf{Activity Datasets}}
		\\
		\cline{2-4}\cline{5-6}
		\multicolumn{1}{l|}{\textbf{\#}}&  \textbf{Opportunity} & \textbf{Skoda} & \textbf{HandGesture} & 
		\textbf{\rev{Utwente}} &
		\textbf{MotionSense} \\ \cline{1-6}
		{0} & null & null& null & --- & ---  \\ \arrayrulecolor{black!10}\cline{2-6}\arrayrulecolor{black!100}
    	  {1} & open door1 & write notes & open window & walking & standing\\ \arrayrulecolor{black!10}\cline{2-6}\arrayrulecolor{black!100}
		  {2} & open door2 & open hood & close window & jogging & stairs-down \\ \arrayrulecolor{black!10}\cline{2-6}\arrayrulecolor{black!100}
		  {3} & close door1 & close hood & water a plant & cycling &  stairs-up\\ \arrayrulecolor{black!10}\cline{2-6}\arrayrulecolor{black!100}
		  {4} & close door2 & check front door & turn book & stairs-up &  walking \\ \arrayrulecolor{black!10}\cline{2-6}\arrayrulecolor{black!100}
		  {5} & open fridge & open left f door & drink  a bottle & stairs-down & jogging \\ \arrayrulecolor{black!10}\cline{2-6}\arrayrulecolor{black!100}
		  {6} & close fridge & close left f door & cut w/ knife & sitting &  \\ \arrayrulecolor{black!10}\cline{2-6}\arrayrulecolor{black!100}
		  {7} & open washer & close left doors & chop w/ knife & standing &--- \\\arrayrulecolor{black!10}\cline{2-6}\arrayrulecolor{black!100}
		  {8} & close washer & check trunk  & stir in a bowl & typing & --- \\\arrayrulecolor{black!10}\cline{2-6}\arrayrulecolor{black!100}
		  {9} & open drawer1 & open/close trunk & forehand & writing & --- \\\arrayrulecolor{black!10}\cline{2-6}\arrayrulecolor{black!100}
		  {10} & close drawer1 & check wheels & backhand & eating & --- \\\arrayrulecolor{black!10}\cline{2-6}\arrayrulecolor{black!100}
		  {11} & open drawer2 & --- & smash & smoking & --- \\\arrayrulecolor{black!10}\cline{2-6}\arrayrulecolor{black!100}
		  {12} & close drawer2 & --- & --- & ---& ---\\\arrayrulecolor{black!10}\cline{2-6}\arrayrulecolor{black!100}
		  {13} & open drawer3 & --- & --- &---& ---\\\arrayrulecolor{black!10}\cline{2-6}\arrayrulecolor{black!100}
		  {14} & close drawer3 & --- & --- &---& ---\\\arrayrulecolor{black!10}\cline{2-6}\arrayrulecolor{black!100}
		  {15} & clean table & --- & --- &---& ---\\\arrayrulecolor{black!10}\cline{2-6}\arrayrulecolor{black!100}
		  {16} & drink cup & --- & --- &---& ---\\\arrayrulecolor{black!10}\cline{2-6}\arrayrulecolor{black!100}
		  {17} & toggle switch & --- & --- & ---&---\\ \cline{1-6}
		\multicolumn{1}{l|}{\textit{$N$}}&  {4} &  {1} & {2} & {6} & {24}\\ \cline{1-6}
		\multicolumn{1}{l|}{$M$}&  {113} &  {57} &  {15} & {9} &  {12} \\ \cline{1-6}
		\multicolumn{1}{l|}{\textit{S.R.}}&  {30 Hz} &  {30 Hz} & {30 Hz} &  {50 Hz} & {50 Hz}\\ \cline{1-6}
		\end{tabular}
		}
\end{table}

\subsection{Replacement} 
\label{sec:rae_eval}

Let the $B$ classes of inference, $I=\{I_1, ...,I_i, ...,I_j,... , I_B\}$, be divided into three categories: (i) {\em Required}, $\mathcal{R}=\{I_{1}, ..., I_{i}\}$, (ii) {\em Sensitive}, $\mathcal{S}=\{I_{i+1}, ..., I_{j}\}$, and (iii) {\em Neutral}, $\mathcal{N}=\{I_{j+1}, ..., I_{B}\}$. \rev{Considering a target app and its potential users, we assume $\mathcal{S}$ is the set of inferences that users wish to keep private. We assume these are sufficiently
{\em sensitive} that the user would wish to prevent the app from making any inferences within this set. Moreover, $\mathcal{R}$ is the set of {\em required} inferences that users
gain utility from if the app can accurately infer them. Finally, $\mathcal{N}$ is the set of neutral inferences that are not sensitive to users that these inferences can be made by the app and it is also not useful for gaining utility. We assume these lists are available to the RAE for its training.}

\subsubsection{Gesture Datasets}
Here, we implemented RAE with the following settings. Seven fully-connected layers with size (number of neurons) $inp =(M\times W)$, $\frac{inp}{2}$, $\frac{inp}{8}$, $\frac{inp}{16}$, $\frac{inp}{8}$, 
$\frac{inp}{2}$,  $out=inp$, respectively (except for the Hand-Gesture dataset with a lower dimensionality that the three middle layers are $\frac{inp}{3}$, $\frac{inp}{4}$, $\frac{inp}{3}$). For all datasets, we consider 1 second time-window, $W=30$. All the experiments are performed on 30 epochs with batch size 128.  The activation function for the output layer is $linear$ and for the input and all of the hidden layers is Scaled Exponential Linear Unit~\cite{klambauer2017self}. In our experiments, to retain the overall structure of the reconstructed data, we set $\mathrm{L_R}$ in Eq.~(\ref{eq:rae}) as the point-wise mean square error function.

To evaluate the privacy loss and utility of the RAE's outcomes, both the raw sensor data and the transformed data are given to a DNN classifier, as an envisioned app, and $F1-score$  are calculated in Table~\ref{cnn_accuracy}, Table~\ref{cnn_accuracy2}, and Table~\ref{cnn_accuracy3}. Here we use  $F1-score$ as evaluation metric because it takes both false positives and false negatives into account.  For RAE, false positives (\ie recognizing $\mathcal{R}$ as $\mathcal{S}$) harm the utility and false negatives (\ie recognizing $\mathcal{S}$ as $\mathcal{R}$ or $\mathcal{N}$) harm privacy. \rev{It should be noted that the classification accuracy metric also shows similar patterns\footnote{ Results available at: \url{https://github.com/mmalekzadeh/replacement-autoencoder}}}.

\begin{table}[t!]
	\centering

	\footnotesize
	\caption{  Gesture recognition results ($F1-score$) by a pre-trained convolutional neural network on the Skoda dataset.  
	}\label{cnn_accuracy}
	\begin{tabular}{clcccclcccc}
		\cline{1-4}\cline{6-10}\cline{1-4}\cline{6-10}
		 \#&\textit{Set of Inferences} & $\mathbf{X}$ & $\mathbf{X'}$ &&\#&Set of Inferences & $\mathbf{X}$ & $\mathbf{X'}$\\ \cline{1-4}\cline{6-10}
		&$\mathcal{R}=\{4,8,9,10\}$ & 97.9  & 96.3 &&&$\mathcal{R}=\{1,4,10\}$ & 97.6 & 95.0 \\
		\arrayrulecolor{black!10}\cline{2-4}\cline{7-10}\arrayrulecolor{black!100}
		1&$\mathcal{S}=\{1,5,6,7\}$& {96.2} & {0.0} &&
		3& $\mathcal{S}=\{2,3,8,9\}$ &  {98.0} &  { 0.0} \\ \arrayrulecolor{black!10}\cline{2-4}\cline{7-10}\arrayrulecolor{black!100}
		&$\mathcal{N}=\{0,2,3\}$& 94.3 & 93.4
		&& &$\mathcal{N}=\{0,5,6,7\}$ &  92.3 & 88.2 
		\\ \cline{1-4}\cline{6-10}\cline{1-4}\cline{6-10}
		
		&$\mathcal{R}=\{2,3,5,6,7,9\}$ & 96.5  & 93.2 && &$\mathcal{R}=\{2,3,5,6,7,9\}$ & 95.8 & 91.1 \\ \arrayrulecolor{black!10}\cline{2-5}\cline{7-10}\arrayrulecolor{black!100}
		2&$\mathcal{S}=\{4,8,10\}$&  {97.9} &  {0.0} && 4&$\mathcal{S}=\{4,8,10\}$ &   {97.4} &  { 0.0}\\ \arrayrulecolor{black!10}\cline{2-4}\cline{7-10}\arrayrulecolor{black!100}
		&$\mathcal{N}=\{0,1\}$& 93.9 & 94.8 && 	&$\mathcal{N}=\{0,1\}$ &  94.3 & 92.4\\ \cline{1-4}\cline{6-10}
	\end{tabular}
\end{table}

\begin{table}[t!]
\parbox{.48\linewidth}{
\centering
	\footnotesize
	\caption{ $F1-score$ for the Hand-Gesture dataset.} \label{cnn_accuracy2}
	\footnotesize
	\begin{tabular}{clccc}
	\hline
	\#& Set of Inferences & $\mathbf{X}$ & $\mathbf{X'}$\\ \hline
		 &$\mathcal{R}=\{1,2,3,4,9,10,11\}$ & 94.1  & 90.1 \\ \arrayrulecolor{black!10}\cline{2-5}\arrayrulecolor{black!100}
		1&$\mathcal{S}=\{5,6,7,8\}$&  {95.7} &  {0.3} \\ \arrayrulecolor{black!10}\cline{2-5}\arrayrulecolor{black!100}
		&$\mathcal{N}=\{0\}$& 95.0 & 96.5 \\ 	\hline

		 &$\mathcal{R}=\{1,3,4,5,6,7\}$ & 95.2  & 90.4 \\ \arrayrulecolor{black!10}\cline{2-5}\arrayrulecolor{black!100}
		2&$\mathcal{S}=\{2,8,9,10,11\}$&  {94.5} &  {0.6}\\ \arrayrulecolor{black!10}\cline{2-5}\arrayrulecolor{black!100}
		&$\mathcal{N}=\{0\}$&  95.0 & 97.5\\ 	\hline

		 &$\mathcal{R}=\{1,3,4,5,6,7,8\}$ & 97.2& 93.3 \\ \arrayrulecolor{black!10}\cline{2-5}\arrayrulecolor{black!100}
		3&$\mathcal{S}=\{2,9,10,11\}$ &   {92.5} &  { 0.7} \\ \arrayrulecolor{black!10}\cline{2-5}\arrayrulecolor{black!100}
		&$\mathcal{N}=\{0\}$ &  95.9 & 97.5  \\
			\hline

		 &$\mathcal{R}=\{2,3,5,6,7,9\}$ & 96.1 & 92.1 \\ \arrayrulecolor{black!10}\cline{2-5}\arrayrulecolor{black!100}
		4&$\mathcal{S}=\{4,8,10\}$ &   {97.0} &  { 0.5} \\ \arrayrulecolor{black!10}\cline{2-5}\arrayrulecolor{black!100}
		&$\mathcal{N}=\{0,1\}$ &  95.7 & 97.6  \\ 
			\hline

	\end{tabular}
    }
\hfill
\parbox{.48\linewidth}{
\centering
\footnotesize
	\caption{ $F1-score$ for the Opportunity dataset.} \label{cnn_accuracy3}
	
	\begin{tabular}{clccc} \hline
		 \# & Set of Inferences& $\mathbf{X}$ & $\mathbf{X'}$\\ \hline 
		 &$\mathcal{R}$=\{9,10,...,17\} & 71.8  & 64.3 \\ \arrayrulecolor{black!10}\cline{2-5}\arrayrulecolor{black!100}
		1& $\mathcal{S}$=\{1,2,...,8\}&  {79.1} &  {0.2} \\ \arrayrulecolor{black!10}\cline{2-5}\arrayrulecolor{black!100}
		&$\mathcal{N}$=\{0\}& 88.9 & 89.7 \\ \hline
		
		 &$\mathcal{R}$=\{1,2,...,8,15,17\} & 76.9 & 75.9 \\ \arrayrulecolor{black!10}\cline{2-5}\arrayrulecolor{black!100}
		2& $\mathcal{S}$=\{9,10,...,14\}&  {71.5} &  {1.3}\\ \arrayrulecolor{black!10}\cline{2-5}\arrayrulecolor{black!100}
		&$\mathcal{N}$=\{0,16\}&  84.4 & 82.1\\ \hline
		
		 &$\mathcal{R}$=\{9,10,...,14,16\} & 74.9& 77.1 \\ \arrayrulecolor{black!10}\cline{2-5}\arrayrulecolor{black!100}
		3 &$\mathcal{S}$=\{1,2,3,4,15,17\} &   {76.2} &  { 0.9} \\ \arrayrulecolor{black!10}\cline{2-5}\arrayrulecolor{black!100}
		&$\mathcal{N}$ =\{0,5,6,7,8\}&  85.0 & 81.6  \\ \hline
		
		&$\mathcal{R}$=\{1,2,...,8,15,17\} & 70.3 & 65.0 \\ \arrayrulecolor{black!10}\cline{2-5}\arrayrulecolor{black!100}
		4 &$\mathcal{S}$=\{9,10,...,14,16\} &   {74.9} &  { 6.3} \\ \arrayrulecolor{black!10}\cline{2-5}\arrayrulecolor{black!100}
		&$\mathcal{N}$=\{0,1\} &  93.7 & 92.9  \\ \hline
	\end{tabular}
	}
\end{table}

\begin{figure}[t!]
\centering
	\includegraphics[scale=0.45]{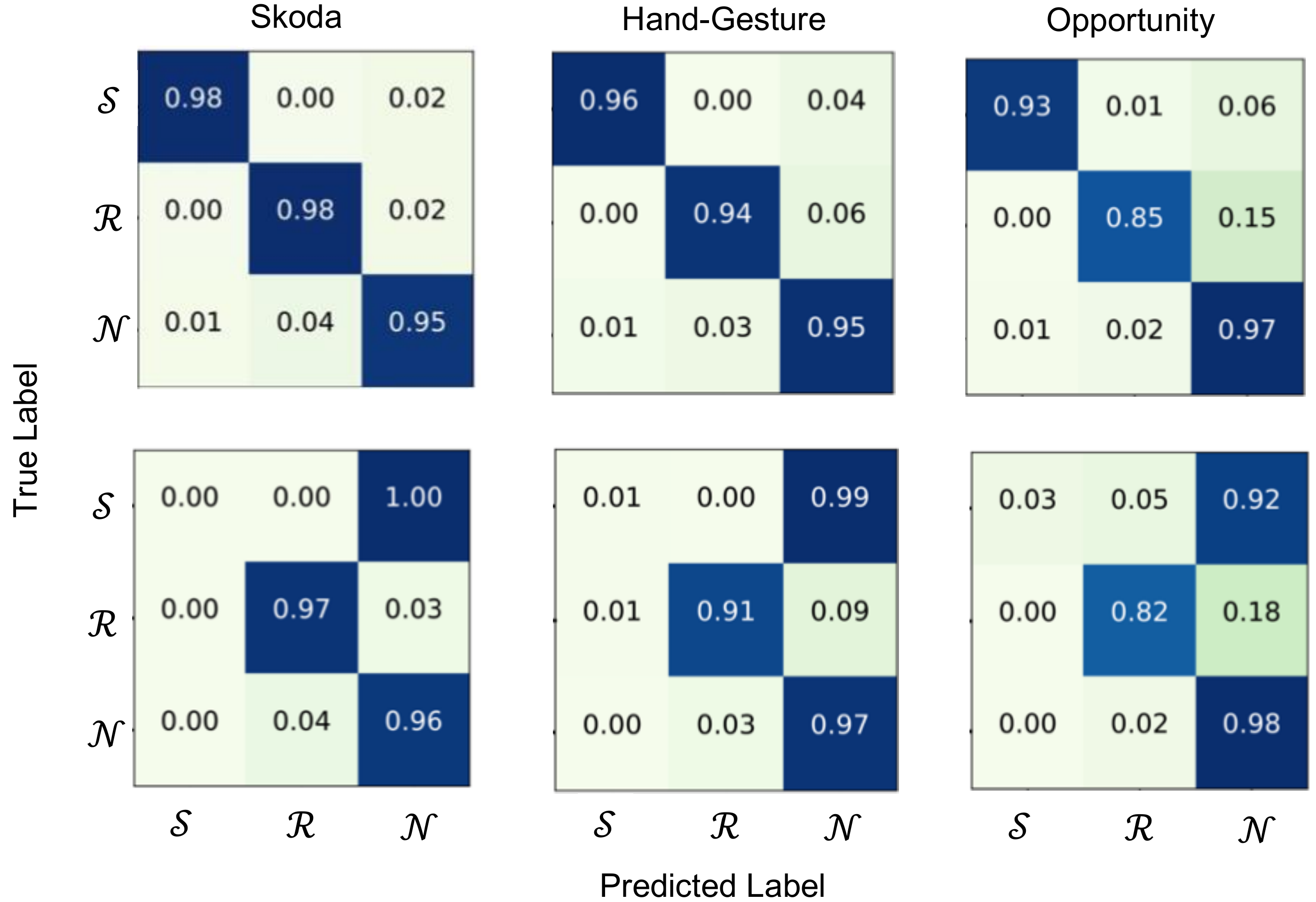}
	\caption{\footnotesize{Confusion Matrix for (top) original  time-series, and (bottom) transformed time-series by RAE. After transformation almost all the {\em sensitive} gestures are recognized as {\em neutral} ones. (Left) Results on the Skoda dataset in Table~\ref{cnn_accuracy} (\#2). (Middle) Hand-Gesture dataset in Table~\ref{cnn_accuracy2} (\#1). (Right) Opportunity dataset in Table~\ref{cnn_accuracy3} (\#2).}}\label{cnf}
\end{figure}

The results show that utility is preserved for non-sensitive, $\mathcal{R}$ and $\mathcal{N}$, classes while recognizing sensitive ones, $\mathcal{S}$, is very unlikely.  Moreover, Figure~\ref{cnf} shows that the model misclassifies all transformed sections corresponding to $\mathcal{S}$ into the $\mathcal{N}$ and therefore the false-positive rate on {\em required} inferences is very low. For instance, to see how RAE can establish a good utility privacy trade-off, consider the results for the Skoda dataset in Table~\ref{cnn_accuracy}(\#2). We see that the gesture classifier can effectively recognize $\mathcal{R}$ gestures (\eg opening and closing doors), even when the app processes the output of RAE instead of the raw data (with 93.2\% accuracy). However, $\mathcal{S}$ gestures (\eg checking doors) that can be recognized with high accuracy when app processes the raw data (with 97.9\% accuracy), are completely filtered out in the output of the RAE. Moreover, the corresponding confusion matrix for this experiment in Figure~\ref{cnf} (Left-Bottom) shows that the utility of the {\em required} inferences is preserved as the classifier wrongly infers all $\mathcal{S}$ gestures as some $\mathcal{N}$ gestures and not as one of $\mathcal{R}$ gestures. 

\subsubsection{\rev{Utwente dataset}}
Let $\mathcal{S}=\{\text{typing, writing, smoking, eating}\}$ be the set of sensitive inferences, $\mathcal{N}=\{\text{sitting still, standing still}\}$ be the set of neutral inferences, and $\mathcal{R}=\{\text{walking, jogging, cycling, stairs-up, stairs-down}\}$ be the set of required inferences. Considering a 2-second time-window ($W=100$), we trained an RAE with 6 hidden layers: 4 Convolutional-LSTM layers using {\em hyperbolic tangent} as the activation function  with 256, 128, 64 and 64 filters respectively, followed by 2 Convolutional layers using Scaled Exponential Linear Unit~\cite{klambauer2017self} as the activation function with 64 and 128 layers respectively. We also put a batch-normalizer on the output of each hidden layer to reduce the training time\footnote{More implementation details and codes for reproducing the  results can be found at \url{https://github.com/mmalekzadeh/replacement-autoencoder}}.

To evaluate the privacy-utility trade-off, we use a DNN classifier. As we see in Table~\ref{cnf_wrist}, the average accuracy of the  classifier on the raw data is more than 99\%. However, when we feed the same classifier with the output of the RAE, all the $\mathcal{S}$ activities are recognized as {\em sitting still}, while the accuracy for $\mathcal{R}$ activities is almost equal to that of the raw data. We observe that for {\em smoking} there still is a 5\% chance of recognition. Note that for some time-windows of the {\em smoking} the raw data are similar to those of {\em standing still}. This effect can be also seen in Table~\ref{cnf_wrist} (column {\em smoking}). We assume this is a labeling error when the data curator labels intervals between cigarette drags as smoking behavior while user is standing.  
\begin{table}[]
\centering
\resizebox{\textwidth}{!}{%
\begin{tabular}{
>{\columncolor[HTML]{FFFFFF}}l 
>{\columncolor[HTML]{FFFFFF}}c 
>{\columncolor[HTML]{FFFFFF}}c 
>{\columncolor[HTML]{FFFFFF}}c 
>{\columncolor[HTML]{FFFFFF}}c 
>{\columncolor[HTML]{FFFFFF}}c 
>{\columncolor[HTML]{FFFFFF}}c 
>{\columncolor[HTML]{FFFFFF}}c 
>{\columncolor[HTML]{FFFFFF}}c 
>{\columncolor[HTML]{FFFFFF}}c 
>{\columncolor[HTML]{FFFFFF}}c 
>{\columncolor[HTML]{FFFFFF}}c }
\multicolumn{1}{c}{\cellcolor[HTML]{FFFFFF}\textbf{}} & {\color[HTML]{036400} \textbf{walking}} & {\color[HTML]{036400} \textbf{jogging}} & {\color[HTML]{036400} \textbf{cycling}} & {\color[HTML]{036400} \textbf{stairs-up}} & {\color[HTML]{036400} \textbf{stairs-down}} & {\color[HTML]{00009B} \textbf{sitting}} & {\color[HTML]{00009B} \textbf{standing}} & {\color[HTML]{CB0000} \textbf{typing}} & {\color[HTML]{CB0000} \textbf{writing}} & {\color[HTML]{CB0000} \textbf{eating}} & {\color[HTML]{CB0000} \textbf{smoking}} \\ \cline{2-12} 
\multicolumn{1}{r|}{\cellcolor[HTML]{FFFFFF}{\color[HTML]{036400} \textbf{walking}}} & \textbf{97.5 $\rightarrow$ 97.2} &  &  & 0.7 $\rightarrow$ 0.7 & 1.5 $\rightarrow$ 1.9 &  &  &  &  &  & 0.3 $\rightarrow$ 0.1 \\
\multicolumn{1}{r|}{\cellcolor[HTML]{FFFFFF}{\color[HTML]{036400} \textbf{jogging}}} &  & \textbf{100 $\rightarrow$ 100} &  &  &  &  &  &  &  &  &  \\
\multicolumn{1}{r|}{\cellcolor[HTML]{FFFFFF}{\color[HTML]{036400} \textbf{cycling}}} &  &  & \textbf{100 $\rightarrow$ 100} &  &  &  &  &  &  &  &  \\
\multicolumn{1}{r|}{\cellcolor[HTML]{FFFFFF}{\color[HTML]{036400} \textbf{stairs-up}}} & 0.4 $\rightarrow$ 0.3 & 0.4 $\rightarrow$ 0.4 & 0.0 $\rightarrow$ 0.1 & \textbf{98.8 $\rightarrow$ 98.8} &  &  & 0.1 $\rightarrow$ 0.1 &  &  &  & 0.3 $\rightarrow$ 0.3 \\
\multicolumn{1}{r|}{\cellcolor[HTML]{FFFFFF}{\color[HTML]{036400} \textbf{stairs-down}}} &  &  &  & 0.3 $\rightarrow$ 0.3 & \textbf{99.7 $\rightarrow$ 99.7} &  &  &  &  &  &  \\
\multicolumn{1}{r|}{\cellcolor[HTML]{FFFFFF}{\color[HTML]{00009B} \textbf{sitting}}} &  &  & 0.0 $\rightarrow$ 0.3 &  &  & \textbf{98.6 $\rightarrow$ 96.8} &  & 1.0 $\rightarrow$ 0.0 & 0.1 $\rightarrow$ 0.0 & 0.1 $\rightarrow$ 0.0 & 0.1 $\rightarrow$ 2.8 \\
\multicolumn{1}{r|}{\cellcolor[HTML]{FFFFFF}{\color[HTML]{00009B} \textbf{standing}}} &  &  & 0.0 $\rightarrow$ 0.3 &  &  &  & \textbf{99.4 $\rightarrow$ 98.2} &  &  &  & 0.6 $\rightarrow$ 1.5 \\
\multicolumn{1}{r|}{\cellcolor[HTML]{FFFFFF}{\color[HTML]{CB0000} \textbf{typing}}} &  &  &  &  &  & \textbf{0.0 $\rightarrow$ 100} &  & \textbf{100 $\rightarrow$ 0.0} &  &  &  \\
\multicolumn{1}{r|}{\cellcolor[HTML]{FFFFFF}{\color[HTML]{CB0000} \textbf{writing}}} &  &  & 0.0 $\rightarrow$ 0.7 &  &  & \textbf{0.0 $\rightarrow$ 99.3} &  &  & \textbf{99.9 $\rightarrow$ 0.0} & 0.1 $\rightarrow$ 0.0 &  \\
\multicolumn{1}{r|}{\cellcolor[HTML]{FFFFFF}{\color[HTML]{CB0000} \textbf{eating}}} &  &  & 0.0 $\rightarrow$ 0.5 &  &  & \textbf{0.1 $\rightarrow$ 99.4} & 0.3 $\rightarrow$ 0.0 &  &  & \textbf{99.6 $\rightarrow$ 0.0} & 0.1 $\rightarrow$ 0.1 \\
\multicolumn{1}{r|}{\cellcolor[HTML]{FFFFFF}{\color[HTML]{CB0000} \textbf{smoking}}} &  &  & 0.0 $\rightarrow$ 0.1 &  &  & \textbf{0.0 $\rightarrow$ 94.9} &  &  &  & 2.3 $\rightarrow$ 0.0 & \textbf{97.5 $\rightarrow$ 5.0}
\end{tabular}%
}
\caption{Confusion Matrix of the results on the test data for Utwente dataset. Rows show the true labels and columns show the predicted labels. In each cell, the left part shows the accuracy on the raw data, and the right part shows the accuracy after transformation. For brevity, all the values are rounded to one decimal point. Empty cells show $0.0 \rightarrow 0.0$.}
\label{cnf_wrist}
\end{table}

\subsection{Anonymization}

\begin{figure}[t!]
\centering
	\includegraphics[scale=0.53]{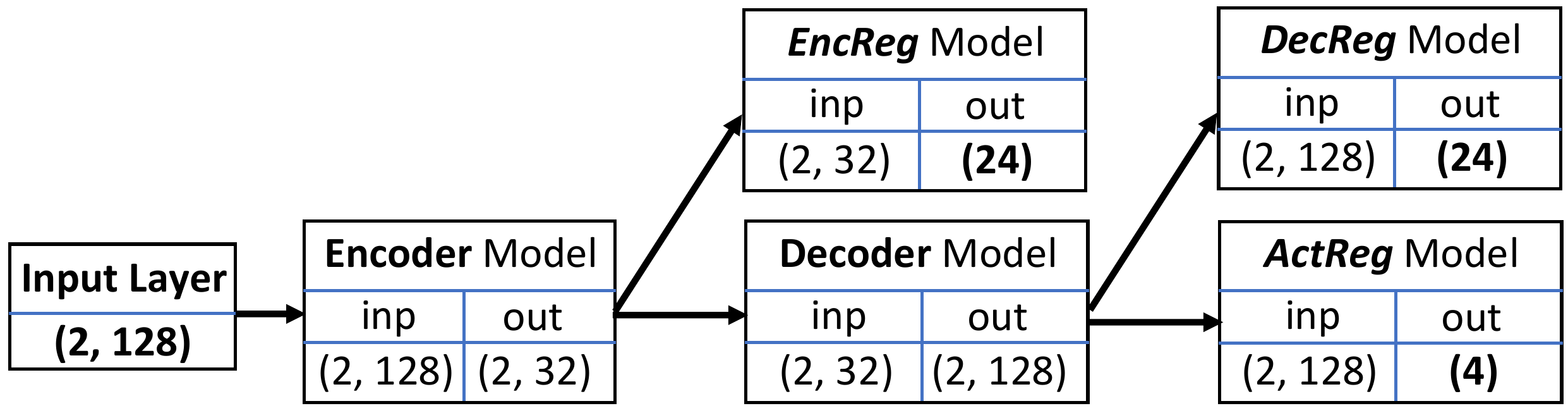}
	\caption{ Details of the training the AAE (Encoder and Decoder) for a  dataset with 24 users and 4 activities. KEY --  \textit{EncReg} and \textit{DecReg}: users' identity recognizers that monitors the output of the Encoder and Decoder, respectively, to reduce the privacy loss; \textit{ActReg}: a users' activity recognizer that monitors the output of the Decoder to increase the utility.}\label{fig:apm2}
\end{figure}

To evaluate the AAE as a data anonymizer, we measure the extent to which the accuracy of activity recognition suffers from anonymization compared to accessing the raw data. We compare the trade-off between recognizing users' activity versus their identity, and compare with baseline methods for coarse-grain time-series data (\textit{resampling} and \textit{singular spectrum analysis}) and with the method in~\cite{edwards2015censoring} that only considers the latent representation by the Encoder model (see Figure~\ref{fig:apm2}), without taking $\mathbf{X}''$ into account.  

We use resampling by Fast Fourier Transform (FFT), which is desirable for periodic time-series, and this is typical with mobile sensor data for activity recognition. Singular Spectrum Analysis (SSA)~\cite{ broomhead1986extracting} is a model-free technique that decomposes time-series into trend, periodic, and structureless (or noise) components using singular value decomposition~(SVD). In our case, we decompose $\mathbf{X}=\{\mathbb{X}^1, \mathbb{X}^2, \ldots, \mathbb{X}^D\}$ such that the $\mathbb{X}^i$ and $\mathbb{X}^{i+1}$  are arranged in descending order according to their corresponding singular value and the original time-series can be recovered as:
%
$
\mathbf{X} = \sum_{i=1}^{D} \mathbb{X}^i.
$
%
Thus, we test the idea of {\em incremental reconstruction} by SSA as a base-line transformation method.

We consider two methods of dividing the dataset into training and test sets, namely {\em Subject} and {\em Trial}.
For {\em Subject}, we put all data of $4$ of the users in the dataset, $2$ females and $2$ males, as testing data and the remaining $20$ users as training. Hence, after training the AAE, we evaluate the model on a dataset of new unseen users.
For {\em Trial}, we put one trial data of each user as testing data and the remaining trials of that user's data as training. For example, where we have three  walking trials for every user, we consider one trial as  testing and the other two as training. In both cases, we put $20\%$ of the training data for validation during the training phase. We repeat each experiment 5 times and report the mean and the standard deviation. For all the experiments we use the magnitude value for both gyroscope and accelerometer. 

To simplify the process of encoding data into a lower-dimensional representation and then decoding it to the original dimension with convolutional filters, we set $W$ to be a power of $2$.  The larger $\mathrm{W}$, the lower the possibility of taking advantage of the correlation among the successive windows by adversaries~\cite{malekzadeh2018mobile}. But larger window sizes increase the delay for real-time apps. We set ${W}=128$ (\ie 2.56 seconds).

For all the regularizers, \textit{EncReg}, \textit{DecReg}, and \textit{ActReg} (see Figure~\ref{fig:apm2}), we use 2D convolutional neural networks. To prevent overfitting, we add a Dropout~\cite{srivastava2014dropout} layer after each convolution layer. We also use L2 regularization to penalize large weights. 
We train the classifier on the original and the anonymized training dataset, and then use it for inference on the test data. We use the Subject setting, thus the test data includes data of new unseen users.

To measure the utility, we train an activity recognition classifier on both the raw data and the output of each transformation method: {\em Resampling}, {\em SSA}, \cite{edwards2015censoring}, and our {\em AAE}. Then, we use the trained model for inference on the corresponding testing data. Here we use the {\em Subject} setting, thus the testing data include data of new unseen users. The second row of Table~\ref{tab:results} (ACT) shows that the average accuracy for activity recognition for both Raw and AAE data is around 92\%. Compared to other methods that decrease the utility of the data, we can preserve the utility and even slightly improve it, on average, as the AAE shapes data such that an activity recognition classifier can learn better from the transformed data than from the raw data. 

\begin{table*}[t]
\centering

\footnotesize
\caption{ Trade-off between utility (activity recognition) and privacy (protecting identity). The forth row shows the K-NN rank between 24 users (the lower the better). Key -- \textit{ACT}: activity recognition, \textit{ID}: identity recognition, \textit{ACC}: accuracy, \textit{F1}:  $F1-score$, \textit{DTW}: Dynamic Time Warping as similarity measure, \textit{SSA}: Singular Spectrum Analysis, \textit{AAE}: Our method. } 
\begin{tabular}{llccccccc}

\hline
 \textit{Experiment} & \textit{Measure} & \multicolumn{1}{c}{\textit{Raw Data}}&
 \multicolumn{2}{c}{{\textit{Resampling}}}
  & \multicolumn{2}{c}{{\textit{SSA}}} & \multicolumn{1}{c}{\cite{edwards2015censoring}} &
 \multicolumn{1}{c}{\textit{\textbf{ AAE}}}  \\
 &&\multicolumn{1}{c}{\textit{50Hz}}&\multicolumn{1}{c}{\textit{10Hz}}&\multicolumn{1}{c}{\textit{5Hz}}&\multicolumn{1}{c}{\textit{1+2}}&\multicolumn{1}{c}{\textit{1}}&\multicolumn{1}{c}{\textit{50Hz}}&\multicolumn{1}{c}{\textit{50Hz}} \\ \hline
\multirow{2}{*}{\begin{tabular}[l]{@{}l@{}} \textit{ACT}\end{tabular}} & mean F1 & 92.5 & 91.1 & 88.0 & 88.6 & 87.4 & 91.5 &  \textbf{92.9} \\ \arrayrulecolor{black!20}\cline{2-9}\arrayrulecolor{black!100}
 & xiance F1 & 2.1 & 0.6 & 1.8 & 0.9 & 0.9 & 0.9 & \textbf{0.37} \\ \hline
\multirow{2}{*}{\begin{tabular}[l]{@{}l@{}} \textit{ID}\end{tabular}} & mean ACC & 96.2 & 31.1 & 13.5 & 34.1 & 16.1 & 15.9 & \textbf{7.0}\\ \arrayrulecolor{black!20}\cline{2-9}\arrayrulecolor{black!100}
 & mean F1 & 95.9 & 25.6 & 8.9 & 28.6 & 12.6  & 11.2 & \textbf{1.8} \\ \hline
\multirow{2}{*}{\begin{tabular}[l]{@{}l@{}} \textit{DTW}\end{tabular}} & mean Rank & 0 & 7.2 & 9.3 & 6.8 & 9.5 & 10.7 & \textbf{6.6} \\ \arrayrulecolor{black!20}\cline{2-9}\arrayrulecolor{black!100}
 & variance Rank & 0 & 5.7 & 5.8 & 5.6 & 5.4 & 5.5 & \textbf{4.7} \\ \hline
\end{tabular}

\label{tab:results}
\end{table*}

To measure the privacy loss, we assume that an adversary has access to the training dataset and we measure the ability of a pre-trained deep classifier on users raw data in inferring the identity of the users when it receives the transformed data. We train a classifier in the {\em Trial} setting over raw data and then feed it different types of transformed data. The third row of
Table~\ref{tab:results} (ID) shows that downsampling data from 50Hz to 5Hz reveals more information than using the AAE output in the original frequency. These results show that the AAE can effectively obscure user-identifiable information so that even a model that has had access to the  original data of the users cannot distinguish them after applying the transformation.
 
Finally, to evaluate the privacy loss and efficiency of the anonymization with an unsupervised mechanism, we implement the $k$-Nearest Neighbors ($k$-NN) with Dynamic Time Warping (DTW)~\cite{salvador2007toward}.
Using DTW, we measure the similarity between the transformed data of a target user ${k}$ and the raw data of each user ${l}$, $\mathbf{X}^{l}$, for all ${l} \in \{{1},\ldots,{k},\ldots,{N}\}$. Then we use this similarity measure to find the ${k}$ nearest neighbors of user ${l}$ and check their rank.
The last row of Table~\ref{tab:results} (DTW) shows that it is very difficult to find similarities between the transformed  and raw data of the users as the performance of the AAE is very similar to the baseline methods and the constraint in Eq.~(\ref{eq:id_act}) maintain the data as similar as possible to the original data. This result shows that the utility-privacy trade-off of AAE is preferable to that of the other methods. 

\subsection{Compound Architecture}\label{sec:evl_comb}
Here, we evaluate a setting where anonymization with the AEE follows replacement using RAE. Considering MotionSense dataset, we want an app to be unable to infer gender or jogging activity  from motion data. Let $\mathcal{S}$=\{{\em jogging}\} be the {\em sensitive} activity to be replaced with $\mathcal{N}$=\{{\em standing still}\} as the {\em neutral} activity. We also consider $\mathcal{R}=$\{{\em walking, stairs-down, stairs-up}\} as the {\em required} inferences. Let the time-window be 2.56 seconds ($W=128$ samples) and $M=2$, \ie ~we consider the magnitude of rotation and acceleration of the device.

First, we train two convolutional neural networks as activity and gender classifiers on the original training dataset. Second, RAE is trained to replace the jogging time-windows while keeping the {\em required} time-windows unmodified in the RAE's output, $\mathbf{X'}$. Third, we use the RAE's output, $\mathbf{X'}$, as the AAE's input and train the AEE to reduce the likelihood of the user's gender being  inferred from the ultimate data that is shared with the app, $\mathbf{X''}$. Finally, after training both autoencoders, we feed the testing dataset into the compound model.   

\begin{table}[t!]
\footnotesize

\centering
\caption{ True-positive rate for each activity and gender classification accuracy ($\%$) using a convolutional neural network for each stage of the compound model on MotionSense~\cite{malekzadeh2018protecting} dataset.}\label{tab:comb}
\begin{tabular}{llcccc} 
\hline
 \multicolumn{2}{l}{\textit{Inference}} & \textit{$\mathbf{X}$: Original } &  \textit{$\mathbf{X'}$: Replacement} & 
\multicolumn{2}{c}{\textit{$\mathbf{X''}$: Anonymization}}\\ \arrayrulecolor{black!100}\cline{5-6}\arrayrulecolor{black!100}
\multicolumn{2}{l}{\textit{}}&\textit{}&\textit{}& $\beta_i = \beta_a = \beta_d$
&
$\beta_i = \frac{1}{2}\beta_a = \beta_d$
 \\ \hline
 & stairs-down & 98.0  & 93.9 & 98.5 & 96.3\\ \arrayrulecolor{black!20}\cline{2-6}\arrayrulecolor{black!100}
$\mathcal{R}$ & stairs-up &  96.4 & 97.8 & 92.3 & 96.3 \\ \arrayrulecolor{black!20}\cline{2-6}\arrayrulecolor{black!100}
 & walking &  99.7 & 94.8 & 89.4 & 94.8 \\ \arrayrulecolor{black!20}\cline{1-6}\arrayrulecolor{black!100}
$\mathcal{S}$ & \textbf{jogging} &  \textbf{99.3} & \textbf{1.4 (92 as $\mathcal{N}$)} & \textbf{.2 (92 as $\mathcal{N}$)} & \textbf{.1 (84 as $\mathcal{N}$)}\\ \arrayrulecolor{black!20}\cline{1-6}\arrayrulecolor{black!100}
$\mathcal{N}$ & standing &  99.9 & 99.9 & 100 & 99.9 \\ \hline
\multicolumn{2}{l}{\textbf{Gender}}   & \textbf{98.9} & \textbf{97.1} &  \textbf{45.0} & \textbf{39.0} \\ \hline
\end{tabular}
\end{table}

Table~\ref{tab:comb} shows the activity and gender  classification results at each processing stage. While $\mathbf{X}$ is highly informative for all inferences,  after replacement \textit{jogging} intervals are not inferred in $\mathbf{X'}$ and they are classified as \textit{standing}. However, \textit{gender} can still be inferred from $\mathbf{X'}$. Inferring gender from $\mathbf{X''}$ (\ie~after anonymization) reaches the desired level of random guess while the inference of $\mathcal{R}$ is maintained close to the original accuracy. Importantly, the proposed framework allows us to give different weights on  preserving the activity and hiding gender: the last column of Table~\ref{tab:comb} shows that a better accuracy can be obtained if we increase the risk of leaking more sensitive information. Notice that, the random guess is 50\% accurate. Thus, the privacy loss is larger when we have 39\% accuracy for gender classifier than 45\%.
 
\section{Discussion}
While we believe the proposed mechanisms establish effective utility-privacy trade-offs for sensor data transformations, here we discuss directions of this work that need more explorations.  

First, in the available datasets, the activities/gestures that are categorized into {\em sensitive}, {\em required}, and {\em neutral} are independent of each other and at each time-window only one of them is happening. However, in the real-world situations there might be correlations among different activities that affect the provided privacy guarantees for some sensitive inferences. Similarly, correlations among consecutive time-windows of a specific activity may incrementally reveal information that facilitate user re-identification. To assess this, we would need access to multi-labelled data collected over a much longer time period as well as a large number of demographically different users. 

Second, to show that the proposed mechanisms can generalize, we performed evaluations on several datasets collected from different type of sensors located in different part of users' body. Current public datasets of mobile and wearable sensor data do not simultaneously satisfy the requirements of abundance and variety of activities and users. To reduce the risk of overfitting, we performed our experiments on DNNs with small number of layers and small number of neurons in each layer. With larger datasets, one can increase the learning capacity of the RAE and AAE by adding more layers to the neural network or investigate various DNN architectures.

Third, we have assumed the existence of a publicly available dataset to train the 
RAE and the AAE. when such public dataset is not available, one option is to use privacy-preserving model training without collecting personal data~\cite{abadi2016deep}, or training the required model through a federated learning~\cite{bonawitz2019towards}. 

Finally, we aim to investigate a privacy-preserving mechanism that transforms sensitive patterns into a mixture of neutral activities rather than only one of them. Moreover, we aim to look for, or to collect, larger datasets to conduct experiments on additional tasks, to derive statistical bounds for the amount of privacy achieved, and to measure the cost of running the proposed local transformations on user devices.

\section{Conclusion}

In this paper we showed how to achieve a trade-off between privacy and utility for sensor data release with an appropriate learning process. In particular, we presented new ways to train deep autoencoders for continuous data transformations to prevent a honest-but-curious app from discovering  users' sensitive information. Our model is general and can be applied to unseen data of new users, without need for re-training.  Experiments conducted on various types of real-world sensor data showed that our transformation mechanism eliminates the possibility of making  sensitive inferences and obscures user-specific motion patterns that enable user re-identification, introducing a small utility loss for activity and gesture recognition tasks.

\section*{Acknowledgment}
The work was supported by the Life Sciences Initiative at Queen Mary University of London and a Microsoft Azure for Research Award (CRM:0740917). Andrea Cavallaro wishes to thank the Alan Turing Institute (EP/N510129/1), which is funded by the EPSRC, for its support through the project PRIMULA. Hamed Haddadi was partially supported by the EPSRC Databox grant (EP/N028260/1). 

\bibliographystyle{elsarticle-num}
\footnotesize

\end{document}